\PassOptionsToPackage{hyphens}{url}
 
\documentclass[11pt]{article}
\usepackage[final]{acl}
 
\usepackage{times}
\usepackage{latexsym}
\usepackage[T1]{fontenc}
\usepackage[utf8]{inputenc}
\usepackage{microtype}
\usepackage{inconsolata}
\usepackage{amsmath,amssymb,amsfonts}   
\usepackage{algorithm}       
\usepackage{algpseudocode}
\usepackage{graphicx}        
\usepackage{textcomp}        
\usepackage{xcolor}          
\usepackage[normalem]{ulem}  
\usepackage{booktabs}        
\usepackage{multirow}        
\usepackage{stfloats}        
\usepackage{enumitem}        
\newcommand{\fakeparagraph}[1]{\noindent\textbf{#1.}}

\setlist[itemize]{leftmargin=*,topsep=2pt,itemsep=1pt,parsep=0pt,partopsep=0pt}
\AtBeginEnvironment{algorithmic}{\footnotesize}
 
\title{PV-Decoding: Progressive Verification Decoding with Decode Gating for Diffusion Language Models}
 
\author{
    \textbf{Renxi Liu\textsuperscript{1,*}},
    \textbf{Yiming Li\textsuperscript{1,*}},
    \textbf{Shengyin Sun\textsuperscript{2}},
    \textbf{Chen Chen\textsuperscript{1}},
    \textbf{Xian Wang\textsuperscript{1}},
    \textbf{Shuai Yang\textsuperscript{1}},\\
    \textbf{Huiling Zhen\textsuperscript{1}},
    \textbf{Xianzhi Yu\textsuperscript{1}},
    \textbf{Yaoyuan Wang\textsuperscript{1}},
    \textbf{Weizhe Lin\textsuperscript{1,\textdagger}},
    \textbf{Xiaosong Li\textsuperscript{1}}
    \\
    \textsuperscript{1}Huawei Technologies \\
    \textsuperscript{2}City University of Hong Kong \\
    \textsuperscript{*}Equal contribution. \\
    \textsuperscript{\textdagger}Corresponding author. Email: \texttt{linweizhe1@huawei.com}
}
\begin{document}
 
\maketitle
 
\begin{abstract}
Discrete diffusion language models (dLLMs) offer an attractive alternative to autoregressive generation by updating multiple tokens in parallel, but their inference efficiency is still limited by the cost of safely committing tentative predictions. In particular, speculative decoding for dLLMs is challenging because speculative tokens within the same diffusion block may depend on one another, making aggressive acceptance prone to error propagation and strict verification computationally expensive. We present \textbf{PV-Decoding}, a confidence-driven speculative decoding framework with progressive verification for block diffusion inference. PV-Decoding partitions masked positions into directly decoded tokens, speculative tokens, and unresolved tokens using two confidence thresholds, and then verifies speculative tokens with progressively stronger policies, including single-batch, dual-batch, and multi-batch verification. This design exposes a practical accuracy--efficiency trade-off: weaker verification minimizes overhead, whereas stronger verification improves acceptance precision and robustness. Experiments on three diffusion language models across GSM8K, MBPP, and HumanEval show that PV-Decoding reduces sequential model forward passes (MFPs) by 8\%--37\% relative to confidence-based decoding, while keeping the largest observed accuracy drop to 1.3\%. On Dream-7B-Instruct/GSM8K, PV-Decoding (1-batch) and PV-Decoding (2-batch) improve throughput by 7.0\%--13.6\% under the same hardware budget, whereas PV-Decoding (multi-batch) is slower under the current runtime. These results suggest that confidence-guided speculation with structured verification is a practical path toward efficient and reliable dLLM inference.
\end{abstract}
 

\section{Introduction}

Discrete diffusion language models (dLLMs) iteratively denoise masked sequences. This generation paradigm enables bidirectional conditioning and multi-token parallel updates within each inference step~\cite{nie2025llada,ye2025dream,wu2025fastdllmtrainingfreeaccelerationdiffusion}. Under block diffusion, one forward pass predicts an entire block of masked positions rather than a single next token. Such parallelism makes dLLMs attractive for low-latency generation, but it also changes the central inference problem: when many positions are updated simultaneously, the decoder must decide which tentative tokens can be safely committed without degrading the conditioning context for later denoising steps.

Speculative decoding is a natural way to further reduce the number of denoising steps, yet verification is substantially harder for dLLMs than for autoregressive models. A speculative token in a diffusion block is typically conditioned on other tentative tokens produced in the same round. As a result, aggressive acceptance may amplify early mistakes, whereas strict verification can require additional branches whose cost quickly offsets the speedup. The key challenge is therefore not merely to speculate more tokens, but to control the accuracy--efficiency trade-off of verification under limited computation.

To address this challenge, we present \textbf{PV-Decoding}, a confidence-driven diffusion decoding framework with progressive verification. PV-Decoding separates \emph{direct decoding} from \emph{speculative filling} according to token confidence, and then verifies speculative tokens with policies of different rigor. 
This design exposes an explicit operating knob: users can choose lighter or stronger verification depending on their compute budget and tolerance for quality degradation.
PV-Decoding is training-free and integrates confidence decoding with verification for mutually conditioned block tokens; unlike lossless methods, it may change the deterministic decoding trajectory.

Our contributions are three-fold:
\begin{itemize}
    \item We formulate confidence-driven speculation for block diffusion decoding, partitioning masked positions into directly decoded, speculative, and unresolved sets through two confidence thresholds.
    \item We propose three progressive verification policies---single-batch, dual-batch, and multi-batch verification---with next-step verification and decode gating that provide increasingly strong safeguards against erroneous speculative commitments while incurring progressively increasing verification cost.
    \item We provide a unified view of speculative decoding for dLLMs in which verification strength, acceptance precision, and runtime overhead can be traded off explicitly across deployment settings. Wall-clock tests under the same hardware budget verify PV-Decoding (1-batch) and PV-Decoding (2-batch) speedups and expose the branch overhead of PV-Decoding (multi-batch).
\end{itemize}


\section{Related Work}

\fakeparagraph{Discrete Diffusion Language Models And Block Decoding Foundations}
Discrete diffusion and masked diffusion models provide the foundation for denoising over token spaces, from structured discrete diffusion processes with absorbing states~\cite{austin2021structured} to modern masked diffusion language modeling objectives~\cite{sahoo2024simple}.
Recent dLLMs build on this foundation and show that discrete denoising can support competitive language generation while exposing richer parallelism than autoregressive decoding~\cite{nie2025llada,ye2025dream,openpangu2025diffusionbase,googledeepmind2025geminidiffusion,inceptionlabs2025mercury}.
Training-free acceleration methods further adapt block diffusion to practical inference by combining block-wise context reuse with confidence-aware decoding~\cite{wu2025fastdllmtrainingfreeaccelerationdiffusion}. 
dParallel introduces certainty-forcing distillation to accelerate masked-token certainty convergence, further improving the parallelism of dLLM decoding~\cite{chen2026dparallel}. 
These works establish the modeling and decoding foundation on which PV-Decoding operates, but they do not directly address how speculative tokens should be verified once multiple positions are updated in parallel.


\fakeparagraph{Speculative Decoding For dLLMs}
Speculative decoding for autoregressive LLMs is commonly formulated as a draft-and-verify procedure: inexpensive proposals from a small model, auxiliary heads, or feature-level predictors are checked by the target model in parallel so that multiple tokens can be accepted without changing the target decoding distribution~\cite{leviathan2023fastinference,chen2023acceleratingllmdecoding,cai2024medusa,li2024eagle}. Earlier and concurrent parallel-decoding systems also validate multiple proposed continuations in one target-model pass, including blockwise parallel decoding with prefix validation~\cite{stern2018blockwise} and tree-based speculative inference and verification~\cite{miao2024specinfer}. In dLLMs, however, speculative verification must account for bidirectional, block-wise conditioning, where tentative updates at different masked positions may mutually influence the verification distribution.
Spiffy~\cite{agrawal2025spiffy} adapts speculative decoding to block diffusion by constructing auto-speculative draft states from the dLLM's own distribution, organizing them with a directed draft graph, and verifying candidate states in parallel while preserving the model's output distribution. FreeDave~\cite{wu2026freedraftandverificationlosslessparallel} similarly frames lossless dLLM acceleration as generating parallel-decoded candidate branches and verifying them so as to reproduce the static decoding trajectory without model modification.
DMax~\cite{chen2026dmax}, Rejection Mixing~\cite{ye2026rejectionmixing}, and WINO~\cite{hong2025wideinnarrowout} are adjacent parallel-decoding approaches because they accept or propagate additional tokens beyond a conservative confidence-based baseline, although their mechanisms are not all framed as speculative verification.
Relative to these methods, PV-Decoding focuses on the verification bottleneck itself and exposes a hierarchy of verification policies, allowing the accuracy--efficiency trade-off to be adjusted according to the available compute budget.

A second line of work improves dLLM inference through more aggressive parallel decoding, rather than through explicit speculative-token verification. 
LoPA~\cite{xu2025lopa}, FDM~\cite{mo2025decodinglargelanguagediffusion} accelerates inference by exploring multiple candidate token-filling orders and selecting the branch with the highest average masked token confidence. 
LocalLeap~\cite{kong2025acceleratingdiffusionllminference}, for example, leverages local determinism propagation to reduce decoding steps. Compared with this line of work, PV-Decoding is less concerned with enlarging the parallel decoding frontier itself and more concerned with how tentative token updates should be verified and committed under different compute budgets.

\section{Preliminaries}
\label{sec:preliminaries}

\fakeparagraph{Discrete Diffusion Language Modeling}
Following discrete diffusion and masked diffusion language modeling~\cite{austin2021structured,sahoo2024simple}, a discrete diffusion language model (dLLM) generates a token sequence by iteratively denoising masked inputs. Let $\mathbf{x}^{(0)}=(x^{(0)}_1,\dots,x^{(0)}_L)$ be the target sequence of length $L$, where each token belongs to a vocabulary $\mathcal{V}$ and $[\texttt{MASK}]$ denotes the mask symbol. A standard forward process progressively corrupts tokens into masks according to
\begin{equation*}
q(\mathbf{x}^{(t)}\mid \mathbf{x}^{(t-1)})=\prod_{i=1}^{L} q(x_i^{(t)}\mid x_i^{(t-1)}),
\end{equation*}
with
\begin{equation*}
\begin{aligned}
q(x_i^{(t)}=[\texttt{MASK}] \mid x_i^{(t-1)})&=\beta_t,\\
q(x_i^{(t)}=x_i^{(t-1)} \mid x_i^{(t-1)})&=1-\beta_t,
\end{aligned}
\end{equation*}
where $\beta_t \in (0,1)$ is the corruption probability at step $t$. The reverse process starts from a masked sequence and predicts a less corrupted state through a parameterized denoising model
\begin{equation*}
p_\theta(\mathbf{x}^{(t-1)}\mid \mathbf{x}^{(t)})=\prod_{i=1}^{L} p_\theta(x_i^{(t-1)}\mid \mathbf{x}^{(t)}).
\end{equation*}
In practical dLLM inference, we follow the block diffusion paradigm of Fast-dLLM~\cite{wu2025fastdllmtrainingfreeaccelerationdiffusion}, where the current model input consists of a decoded prefix $\mathbf{x}_0$ and a partially masked block $\mathbf{x}_b=(x_{b,1},\dots,x_{b,B})$ of length $B$. Let $\mathcal{M} \subseteq \{1,\dots,B\}$ denote the masked positions in the current block. A single forward pass of the denoising model produces a token distribution for every masked position:
\begin{equation}
    p_i(v)=p_\theta(x_{b,i}=v \mid [\mathbf{x}_0;\mathbf{x}_b]), \quad i \in \mathcal{M},\, v \in \mathcal{V}.
\end{equation}
We use
\begin{equation}
    c_i=\max_{v \in \mathcal{V}} p_i(v), \qquad
    \hat{x}_{b,i}=\arg\max_{v \in \mathcal{V}} p_i(v),
\end{equation}
to denote the confidence and argmax prediction at position $i$, respectively. Thus, block diffusion can be viewed as the inference-time realization of the dLLM's parallel denoising capability, and these quantities form the basic interfaces shared by the baseline decoder and our method.

\fakeparagraph{Confidence Decoding Baseline}
A standard confidence-based block decoder commits all masked positions whose confidence exceeds a decoding threshold $\tau_{\mathrm{dec}}$. Formally, the set of directly decodable positions is
\begin{equation}
    \mathcal{D}=\left\{ i \in \mathcal{M} \;\middle|\; c_i \ge \tau_{\mathrm{dec}} \right\}.
\end{equation}
For every $i \in \mathcal{D}$, the decoder commits $\hat{x}_{b,i}$ in parallel. If $\mathcal{D}=\emptyset$, the decoder falls back to greedy decoding and commits only the most confident masked position,
\begin{equation}
    i^\star=\arg\max_{i \in \mathcal{M}} c_i.
\end{equation}
This baseline is important for two reasons. First, it is the practical decoding rule used by existing block-diffusion systems. Second, PV-Decoding is defined as an extension of this baseline: instead of committing only high-confidence tokens, it also tentatively fills moderate-confidence tokens and then verifies them in the next step.

\section{Methodology}

\begin{figure*}[t]
    \centering
    \includegraphics[width=0.93\textwidth]{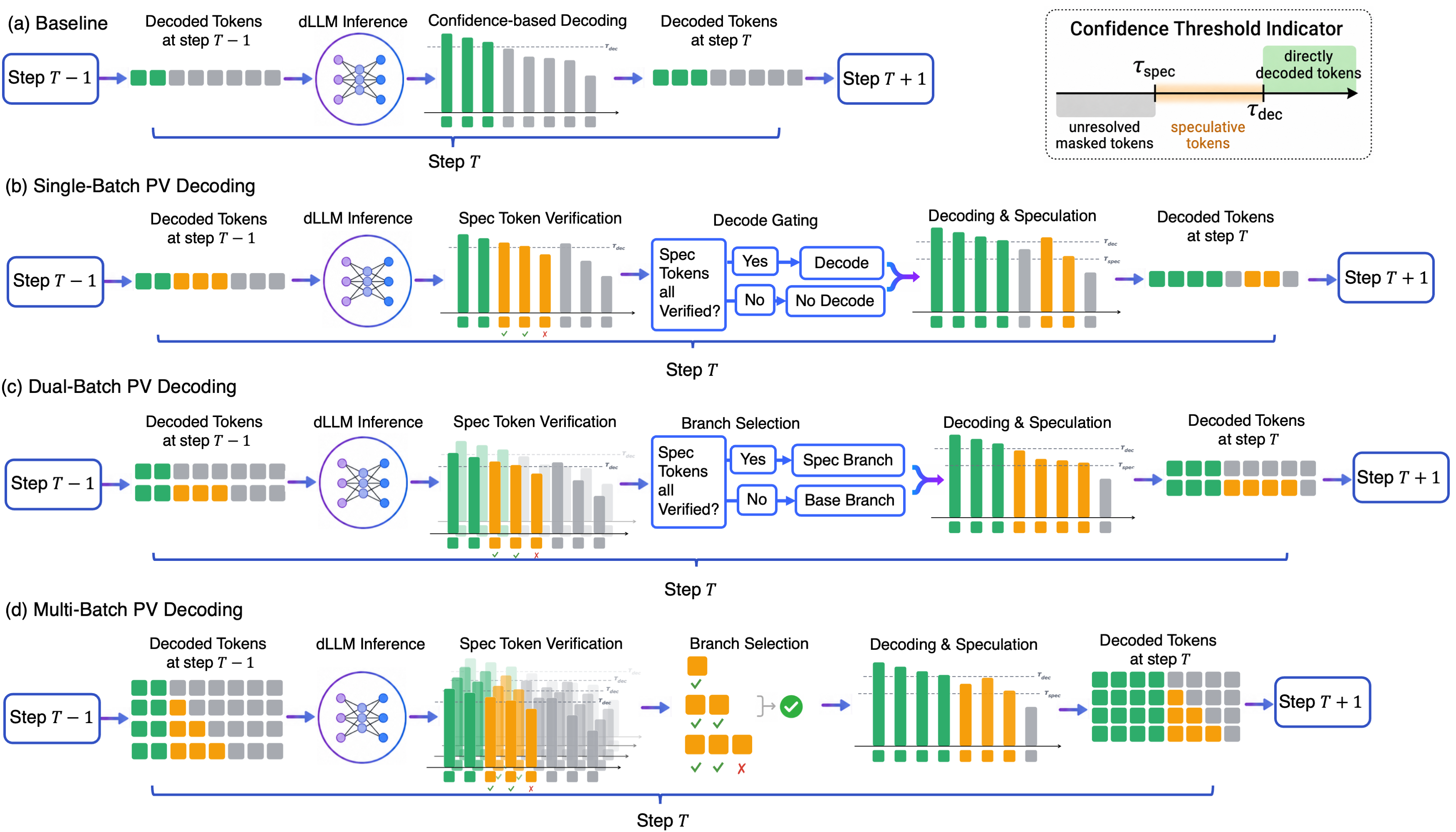}
    \caption{Overview of PV-Decoding under block diffusion decoding. Given an input sequence with partially masked tokens, one forward pass produces token predictions and confidence scores. Two thresholds, $\tau_{\mathrm{spec}}$ and $\tau_{\mathrm{dec}}$, partition masked positions by confidence $c_i$ into unresolved tokens ($c_i < \tau_{\mathrm{spec}}$), speculative tokens ($\tau_{\mathrm{spec}} \leq c_i < \tau_{\mathrm{dec}}$), and directly decoded tokens ($c_i \geq \tau_{\mathrm{dec}}$). The speculative tokens are then processed by progressively stronger verification policies: single-batch, dual-batch, and multi-batch verification.}
    \label{fig:pipeline}
\end{figure*}

\subsection{Problem Formulation}

PV-Decoding is designed for block diffusion decoding, where one forward pass updates multiple masked positions at once. This parallelism is beneficial for efficiency, but it also introduces a verification challenge that does not arise in standard autoregressive decoding. In particular, a speculative token may be conditioned on other tentative tokens proposed in the same round. If such tokens are committed too aggressively, errors can contaminate the context used for subsequent decoding steps, and stronger verification mechanisms incur heavier verification cost.
Observing that different types of tasks entail varying requirements for inference accuracy, we propose PV-Decoding, a method that couples progressive multi-level verification with corresponding verification computation cost.

Figure~\ref{fig:pipeline} summarizes the overall idea. Given a partially masked block, one model pass produces confidence scores for all masked positions. PV-Decoding then partitions them into three groups: directly decoded tokens, speculative tokens, and unresolved masked tokens. Directly decoded tokens are committed immediately, while speculative tokens are only tentatively filled and must be verified in the next iteration. The role of verification is therefore to decide how much speculative context can be trusted before more tokens are committed.

This view motivates a two-stage design. First, PV-Decoding performs \emph{confidence-based speculation}, which enlarges the set of candidate updates beyond the baseline confidence decoder. Second, it applies a \emph{verification policy} that controls how conservatively these tentative updates are accepted. Stronger policies improve acceptance precision but consume more verification compute.

Tentative tokens may affect later predictions before verification; hence PV-Decoding is deterministic but neither trajectory- nor distribution-preserving.

\subsection{Confidence-Based Speculation}

Building on the notation in Section~\ref{sec:preliminaries}, PV-Decoding introduces a second threshold $\tau_{\mathrm{spec}} < \tau_{\mathrm{dec}}$. At decoding step $r$, each masked position $i \in \mathcal{M}^{(r)}$ is assigned to one of three sets:
\begin{equation}
    \mathcal{D}^{(r)} = \left\{ i \in \mathcal{M}^{(r)} \;\middle|\; c_i^{(r)} \ge \tau_{\mathrm{dec}} \right\},
\end{equation}
\begin{equation}
    \mathcal{S}^{(r)} = \left\{ i \in \mathcal{M}^{(r)} \;\middle|\; \tau_{\mathrm{spec}} \le c_i^{(r)} < \tau_{\mathrm{dec}} \right\},
\end{equation}
\begin{equation}
    \mathcal{U}^{(r)} = \mathcal{M}^{(r)} \setminus \left(\mathcal{D}^{(r)} \cup \mathcal{S}^{(r)}\right),
\end{equation}
where $\mathcal{D}^{(r)}$, $\mathcal{S}^{(r)}$, and $\mathcal{U}^{(r)}$ denote the directly decoded, speculative, and unresolved positions, respectively.

For every $i \in \mathcal{D}^{(r)} \cup \mathcal{S}^{(r)}$, PV-Decoding tentatively uses the argmax prediction $\hat{x}_{b,i}^{(r)}=\arg\max_v p_i^{(r)}(v)$. The distinction is in commitment: positions in $\mathcal{D}^{(r)}$ are committed immediately, whereas positions in $\mathcal{S}^{(r)}$ are only inserted as speculative placeholders for the next verification step. As illustrated in Figure~\ref{fig:pipeline}, this rule separates easy predictions from moderate-confidence predictions while leaving low-confidence positions masked.

\subsection{Progressive Verification Policy}

After speculation, PV-Decoding verifies the tokens proposed in the previous iteration. Let $\mathcal{S}_{\mathrm{prev}}$ denote the previously speculated positions and let $\hat{x}^{\mathrm{prev}}_{b,i}$ be the token proposed for position $i \in \mathcal{S}_{\mathrm{prev}}$. A verification policy accepts a subset of these positions only when the current forward pass predicts the same token with confidence at least $\tau_{\mathrm{dec}}$. 
We refer to these verified and committed positions as \emph{accepted speculative tokens}; they are distinct from directly decoded tokens, which cross $\tau_{\mathrm{dec}}$ before speculation, and from unaccepted speculative tokens, which are re-masked. 
The difference between our three variants lies in how much speculative context is used during this check.

\fakeparagraph{Single-Batch Verification}
Single-batch verification uses only the standard decoding pass and therefore has the lowest overhead. All speculative tokens are checked jointly in one branch. A speculative token is accepted when the current argmax matches the previous proposal and its confidence exceeds $\tau_{\mathrm{dec}}$; otherwise, that position is remasked. Importantly, PV-Decoding commits new high-confidence unresolved positions in $\mathcal{U}^{(r)}$ only when \emph{all} speculative tokens from the previous step are verified. This verification-gated rule prevents the decoder from committing fresh tokens on top of a partially incorrect speculative context. Its effective verification batch size is $1$.

\fakeparagraph{Dual-Batch Verification}
Dual-batch verification evaluates two branches in parallel: a \emph{baseline branch} without speculative tokens and a \emph{speculative branch} with all speculative tokens filled in. The speculative branch is accepted only if every speculative token passes token-consistency and confidence checks; otherwise, decoding falls back to the baseline branch. Relative to single-batch verification, this policy improves acceptance precision because the fallback branch is guaranteed to remain speculation-free, while the verification cost increases to an effective verification batch size of $2$.

\fakeparagraph{Multi-Batch Verification}
Multi-batch verification further strengthens conditioning by verifying speculative tokens over a subset graph. Following the subset construction of Spiffy~\cite{agrawal2025spiffy}, each graph node represents a subset of the previously speculative positions. Let $\mathcal{G}=(\mathcal{V},\mathcal{E})$ be this directed acyclic graph (DAG), where each node $v\in\mathcal{V}$ stores a subset $\mathcal{C}(v)\subseteq\mathcal{S}_{\mathrm{prev}}$ and an edge $u\rightarrow v$ indicates $\mathcal{C}(u)\subset\mathcal{C}(v)$. The root is the empty subset, $\mathcal{C}(v_0)=\emptyset$, which corresponds to the baseline branch with all previous speculative positions masked. For a verification budget of $L+1$ branches, PV-Decoding selects at most $L$ nested non-root nodes from this graph, fills the corresponding speculative tokens in each branch, and evaluates all selected branches in one verification batch. A node is verified only if all speculative tokens in the preceding branch have already been accepted and the current branch verifies the same token with confidence above the decoding threshold $\tau_{\mathrm{dec}}$. Among verified nodes, PV-Decoding chooses the node with the largest accepted subset, using cardinality as the primary key and the graph order as a deterministic tie-breaker; this branch becomes the new baseline context, while speculative positions outside the chosen subset are re-masked. Thus $L$ controls the number of subset nodes tested, and the effective verification batch size is $L+1$ including the root baseline branch. Unlike Spiffy, PV-Decoding allows the number of speculative tokens to vary adaptively through $\tau_{\mathrm{spec}}$ and is not tied to a particular DAG construction; we use Spiffy's subset DAG here for convenience, while alternative DAG-selection strategies can also be used.

\fakeparagraph{Unified Perspective}
The three policies therefore form a clear accuracy--efficiency spectrum: single-batch verification has the lowest overhead but the weakest conditioning; dual-batch verification adds one extra branch to reduce spurious acceptance; multi-batch verification applies the strongest conditioning at the highest verification cost. In all cases, speculation is still driven by the same confidence rule, while verification determines how aggressively tentative updates can be committed.

\begin{algorithm}[tbp]
\small
\caption{Unified PV-Decoding}
\label{alg:pv_decoding_unified}
\begin{algorithmic}[1]
\Require Diffusion LLM $f_\theta$, prefix $\mathbf{x}_0$, initial block $\mathbf{x}_b^{(0)}$
\Require Decode threshold $\tau_{\mathrm{dec}}$, speculation threshold $\tau_{\mathrm{spec}}$, verification policy $\pi$
\Ensure Fully decoded block $\mathbf{x}_b^{*}$
\State $\mathbf{x}_b \leftarrow \mathbf{x}_b^{(0)}$
\State $\mathcal{S}_{\mathrm{prev}} \leftarrow \emptyset$; $\{\hat{x}_{b,i}^{\mathrm{prev}}\} \leftarrow \varnothing$
\While{$\mathbf{x}_b$ still contains \textsc{[mask]}}
    \State Verify previous speculative tokens before selecting any new direct or speculative positions.
    \If{$\mathcal{S}_{\mathrm{prev}} \neq \emptyset$}
        \State $(\mathbf{x}_b,\alpha,\mathbf{p}) \leftarrow \textsc{VerifyPolicy}(\pi,\mathbf{x}_0,\mathbf{x}_b,$
        \Statex \hspace{\algorithmicindent}$\mathcal{S}_{\mathrm{prev}},\{\hat{x}_{b,i}^{\mathrm{prev}}\},\tau_{\mathrm{dec}})$
        \State $\mathcal{S}_{\mathrm{prev}} \leftarrow \emptyset$; $\{\hat{x}_{b,i}^{\mathrm{prev}}\} \leftarrow \varnothing$
        \If{$\alpha=\textsc{false}$}
            \State \textbf{continue}
        \EndIf
    \Else
        \State $\mathbf{p} \leftarrow f_\theta([\mathbf{x}_0;\mathbf{x}_b])$
    \EndIf
    \State $\mathcal{S}_{\mathrm{next}} \leftarrow \emptyset$; $\{\hat{x}_{b,i}^{\mathrm{next}}\} \leftarrow \varnothing$
    \For{each masked position $i$ in $\mathbf{x}_b$}
        \State $c_i \leftarrow \max_v p_i(v)$; $\hat{x}_{b,i} \leftarrow \arg\max_v p_i(v)$
        \If{$c_i \ge \tau_{\mathrm{dec}}$}
            \State commit $\hat{x}_{b,i}$ into $\mathbf{x}_b$
        \ElsIf{$\tau_{\mathrm{spec}} \le c_i < \tau_{\mathrm{dec}}$}
            \State tentatively fill $\hat{x}_{b,i}$ and add $i$ to $\mathcal{S}_{\mathrm{next}}$
            \State store $\hat{x}_{b,i}$ in $\{\hat{x}_{b,i}^{\mathrm{next}}\}$
        \EndIf
    \EndFor
    \State $\mathcal{S}_{\mathrm{prev}} \leftarrow \mathcal{S}_{\mathrm{next}}$; $\{\hat{x}_{b,i}^{\mathrm{prev}}\} \leftarrow \{\hat{x}_{b,i}^{\mathrm{next}}\}$
\EndWhile
\State \Return $\mathbf{x}_b^{*} \leftarrow \mathbf{x}_b$
\end{algorithmic}
\end{algorithm}

\begin{algorithm}[tbp]
\small
\caption{Progressive Verification Policy}
\label{alg:verification_policy}
\begin{algorithmic}[1]
\Require Policy type $\pi \in \{\texttt{single},\texttt{dual},\texttt{multi}\}$, prefix $\mathbf{x}_0$, block $\mathbf{x}_b$
\Require Speculative positions $\mathcal{S}_{\mathrm{prev}}$, tokens $\{\hat{x}_{b,i}^{\mathrm{prev}}\}$, decode threshold $\tau_{\mathrm{dec}}$
\Ensure Updated block $\mathbf{x}_b$, new-commit flag $\alpha$, and branch distribution $\mathbf{p}$
\State Return the branch distribution used after verification so the decoder does not reuse stale pre-verification probabilities.
\If{$\pi = \texttt{single}$}
    \State $\mathbf{p} \leftarrow f_\theta([\mathbf{x}_0;\mathbf{x}_b])$
    \State accept $i$ iff $\arg\max_v p_i(v)=\hat{x}_{b,i}^{\mathrm{prev}}$ and $p_i(\hat{x}_{b,i}^{\mathrm{prev}})\ge\tau_{\mathrm{dec}}$
    \State re-mask rejected positions; set $\alpha$ to whether all positions were accepted
\ElsIf{$\pi = \texttt{dual}$}
    \State evaluate a baseline branch $\mathbf{p}^{(0)}$ and a speculative branch $\mathbf{p}^{(1)}$
    \If{all speculative tokens pass consistency and confidence checks in $\mathbf{p}^{(1)}$}
        \State keep the speculative branch and set $\mathbf{p}\leftarrow\mathbf{p}^{(1)}$
    \Else
        \State fall back to the baseline branch, re-mask speculative positions, and set $\mathbf{p}\leftarrow\mathbf{p}^{(0)}$
    \EndIf
    \State $\alpha \leftarrow \textsc{true}$
\Else
    \State build a subset DAG $\mathcal{G}$ over $\mathcal{S}_{\mathrm{prev}}$ with root subset $\emptyset$
    \State select up to $L$ non-root nodes $\{v_\ell\}_{\ell=1}^{L}$ from $\mathcal{G}$
    \State evaluate one verification batch containing the root branch and branches with $\mathcal{C}(v_\ell)$ filled
    \State mark the root verified by default
    \State mark $v_\ell$ verified iff all tokens in $\mathcal{C}(v_\ell)$ pass consistency and confidence checks
    \State choose the verified node $v^*$ with largest $|\mathcal{C}(v^*)|$, using graph order to break ties
    \State accept $\mathcal{C}(v^*)$, re-mask $\mathcal{S}_{\mathrm{prev}}\setminus\mathcal{C}(v^*)$, and set $\mathbf{p}$ to the distribution of branch $v^*$
    \State $\alpha \leftarrow \textsc{true}$
\EndIf
\State \Return $(\mathbf{x}_b,\alpha,\mathbf{p})$
\end{algorithmic}
\end{algorithm}

\section{Experiments}

\subsection{Experimental Settings}

For brevity, we refer to PV-Decoding (1-batch), PV-Decoding (2-batch), and PV-Decoding (multi-batch) as PV-1, PV-2, and PV-multi, respectively.

All experiments are conducted on Huawei Ascend 910B3 NPU hardware. We evaluate PV-Decoding on three diffusion language models: LLaDA-8B-Instruct~\cite{nie2025llada}, Dream-7B-Instruct~\cite{ye2025dream}, and openPangu-7B-Diffusion-Base~\cite{openpangu2025diffusionbase}. Evaluation is performed on GSM8K~\cite{cobbe2021gsm8k}, MBPP~\cite{austin2021program}, and HumanEval~\cite{chen2021evaluating}, which together cover mathematical reasoning and code generation.

Following Fast-dLLM~\cite{wu2025fastdllmtrainingfreeaccelerationdiffusion}, we use block diffusion with KV-cache reuse. 
The diffusion block size is set to $32$ tokens. All evaluations are implemented using the Language Model Evaluation Harness (\texttt{lm\_eval})~\cite{eval-harness}. 
We use a batch size of 1 for accuracy and model forward pass (MFP) evaluation, while wall-clock performance is measured under varying levels of concurrency.
The baseline decoding threshold is fixed at $\tau_{\mathrm{dec}}=0.9$ in accordance with Fast-dLLM, and the maximum number of generated tokens is set to $512$. For PV-Decoding, the default speculative threshold is $\tau_{\mathrm{spec}}=0.8$. For multi-batch verification, the maximum number of verification branches is capped at $8$.
This threshold pair is used throughout without task-specific tuning.

We compare PV-Decoding against Fast-dLLM and representative accelerated decoding baselines, including LoPA~\cite{xu2025lopa}, Spiffy~\cite{agrawal2025spiffy}, and LocalLeap~\cite{kong2025acceleratingdiffusionllminference}. LoPA uses $7$ branches and Spiffy $8$. 
Cross-method comparisons use MFPs as the efficiency metric. Table 2 separately reports hardware-matched wall-clock results for Fast-dLLM and PV-Decoding.

\subsection{Main Results}

\begin{figure*}[t]
    \centering
    \includegraphics[width=0.93\linewidth]{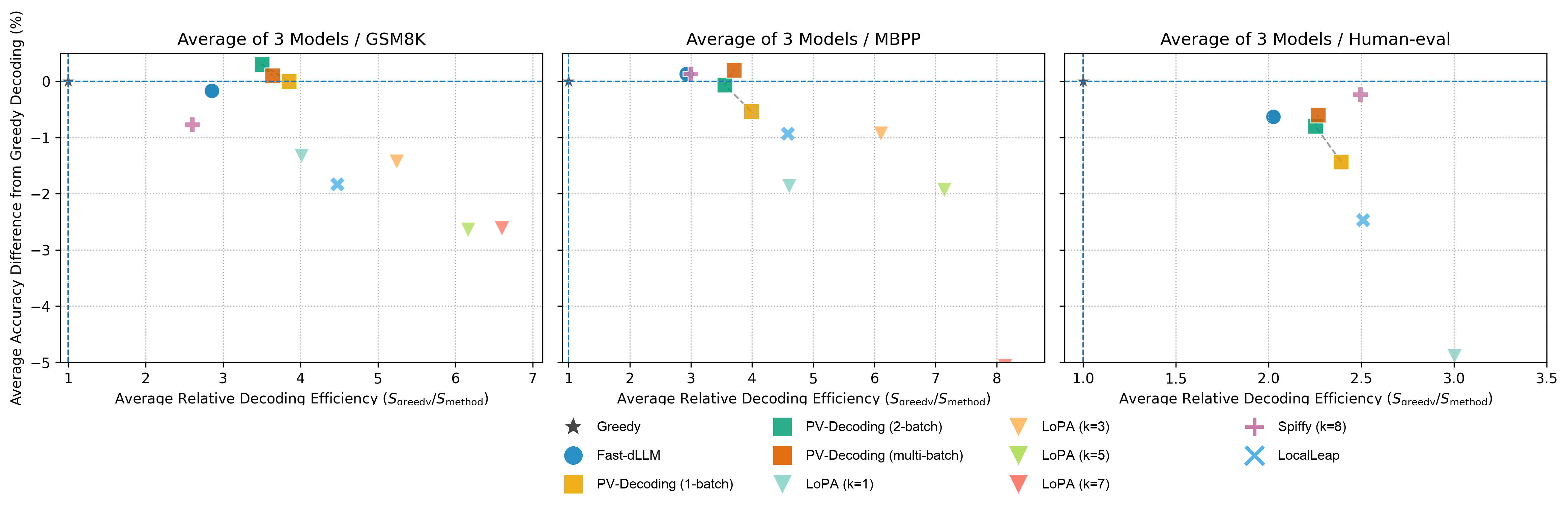}
    \caption{Model-averaged accuracy--efficiency trade-off across benchmarks. For each benchmark and method, relative decoding efficiency and $\Delta$Acc are averaged over LLaDA-8B-Instruct, Dream-7B-Instruct, and openPangu-7B-Diffusion-Base. The x-axis shows the relative decoding efficiency, computed as the ratio between the number of sequential model forward passes (MFPs) required by greedy decoding and that required by each method. The y-axis reports the accuracy difference with respect to greedy decoding.}
    \label{fig:pareto-scatter}
\end{figure*}

The overall accuracy--efficiency trade-off is summarized in Figure~\ref{fig:pareto-scatter}. 
To emphasize the general trend rather than model-specific variance, each point aggregates the corresponding benchmark--method result over the three evaluated diffusion language models.
The x-axis reports relative decoding efficiency, defined as the ratio between the number of greedy MFPs and that required by each method, i.e., $S_{\mathrm{greedy}}/S_{\mathrm{method}}$.
The y-axis reports $\Delta$Acc relative to greedy decoding, so positive values indicate accuracy gains over greedy decoding and negative values indicate accuracy drops. 
Hence, the upper-right region indicates a favorable accuracy--efficiency trade-off. 
The efficiency reported here is measured in terms of sequential model forward passes. Achieving the optimal efficiency of the 2-branch and multi-branch variants would require either optimized operators or a proportional increase in hardware resources.

Table~\ref{tab:main-acc-efp} reports the underlying model-specific accuracy and sequential MFP measurements behind the model-averaged Pareto view.

\begin{table*}[b]
    \centering
    \caption{Main accuracy and model-forward-pass results across diffusion language models and benchmarks. Accuracy is reported in percentage, and MFP denotes the average number of model forward passes. Parenthetical changes of the accuracy and MFP in the PV-Decoding columns are reported relative to Fast-dLLM.}
    \label{tab:main-acc-efp}
    \resizebox{\textwidth}{!}{%
    \begin{tabular}{llcccccccccc}
        \toprule
        \multirow{2}{*}{Model} & \multirow{2}{*}{Dataset}
        & \multicolumn{2}{c}{Greedy}
        & \multicolumn{2}{c}{Fast-dLLM}
        & \multicolumn{2}{c}{\shortstack{PV-Decoding\\(1-batch)}}
        & \multicolumn{2}{c}{\shortstack{PV-Decoding\\(2-batch)}}
        & \multicolumn{2}{c}{\shortstack{PV-Decoding\\(multi-batch)}} \\
        \cmidrule(lr){3-4}\cmidrule(lr){5-6}\cmidrule(lr){7-8}\cmidrule(lr){9-10}\cmidrule(lr){11-12}
        & & Acc (\%) & MFP & Acc (\%) & MFP & Acc (\%) & MFP & Acc (\%) & MFP & Acc (\%) & MFP \\
        \midrule
        \multirow{3}{*}{Dream-7B-Instruct}
        & GSM8K & 72.4 & 122.4 & 72.4 & 58.3 & 72.4 \textcolor{green!50!black}{(+0.0)} & 46.2 \textcolor{green!50!black}{(-12.1)} & 72.9 \textcolor{green!50!black}{(+0.5)} & 52.3 \textcolor{green!50!black}{(-6.0)} & 72.3 \textcolor{red}{(-0.1)} & 46.5 \textcolor{green!50!black}{(-11.8)} \\
        & MBPP & 55.4 & 256.0 & 55.4 & 62.9 & 54.8 \textcolor{red}{(-0.6)} & 42.7 \textcolor{green!50!black}{(-20.2)} & 55.4 \textcolor{green!50!black}{(+0.0)} & 47.2 \textcolor{green!50!black}{(-15.7)} & 55.4 \textcolor{green!50!black}{(+0.0)} & 45.6 \textcolor{green!50!black}{(-17.3)} \\
        & HumanEval & 56.1 & 173.6 & 54.8 & 86.4 & 54.3 \textcolor{red}{(-0.5)} & 75.5 \textcolor{green!50!black}{(-10.9)} & 54.3 \textcolor{red}{(-0.5)} & 78.8 \textcolor{green!50!black}{(-7.6)} & 54.9 \textcolor{green!50!black}{(+0.1)} & 78.2 \textcolor{green!50!black}{(-8.2)} \\
        \midrule
        \multirow{3}{*}{LLaDA-8B-Instruct}
        & GSM8K & 78.5 & 210.0 & 78.2 & 77.1 & 78.2 \textcolor{green!50!black}{(+0.0)} & 70.9 \textcolor{green!50!black}{(-6.2)} & 78.5 \textcolor{green!50!black}{(+0.3)} & 62.2 \textcolor{green!50!black}{(-14.9)} & 78.8 \textcolor{green!50!black}{(+0.6)} & 70.7 \textcolor{green!50!black}{(-6.4)} \\
        & MBPP & 36.2 & 56.0 & 36.6 & 33.2 & 36.4 \textcolor{red}{(-0.2)} & 27.6 \textcolor{green!50!black}{(-5.6)} & 36.4 \textcolor{red}{(-0.2)} & 28.1 \textcolor{green!50!black}{(-5.1)} & 36.8 \textcolor{green!50!black}{(+0.2)} & 28.4 \textcolor{green!50!black}{(-4.8)} \\
        & HumanEval & 37.2 & 160.0 & 36.6 & 74.9 & 36.0 \textcolor{red}{(-0.6)} & 63.5 \textcolor{green!50!black}{(-11.4)} & 36.6 \textcolor{green!50!black}{(+0.0)} & 64.5 \textcolor{green!50!black}{(-10.4)} & 36.6 \textcolor{green!50!black}{(+0.0)} & 64.6 \textcolor{green!50!black}{(-10.3)} \\
        \midrule
        \multirow{3}{*}{openPangu-7B-Diffusion-Base}
        & GSM8K & 77.8 & 407.1 & 77.6 & 108.8 & 78.1 \textcolor{green!50!black}{(+0.5)} & 68.5 \textcolor{green!50!black}{(-40.3)} & 78.2 \textcolor{green!50!black}{(+0.6)} & 85.0 \textcolor{green!50!black}{(-23.8)} & 77.9 \textcolor{green!50!black}{(+0.3)} & 76.6 \textcolor{green!50!black}{(-32.2)} \\
        & MBPP & 51.4 & 511.0 & 51.4 & 169.0 & 50.2 \textcolor{red}{(-1.2)} & 129.9 \textcolor{green!50!black}{(-39.1)} & 51.0 \textcolor{red}{(-0.4)} & 157.7 \textcolor{green!50!black}{(-11.3)} & 51.4 \textcolor{green!50!black}{(+0.0)} & 144.7 \textcolor{green!50!black}{(-24.3)} \\
        & HumanEval & 47.0 & 141.6 & 47.0 & 73.4 & 45.7 \textcolor{red}{(-1.3)} & 60.1 \textcolor{green!50!black}{(-13.3)} & 47.0 \textcolor{green!50!black}{(+0.0)} & 68.3 \textcolor{green!50!black}{(-5.1)} & 47.0 \textcolor{green!50!black}{(+0.0)} & 67.3 \textcolor{green!50!black}{(-6.1)} \\
        \bottomrule
    \end{tabular}%
    }
\end{table*}

Compared with both greedy decoding and confidence-based decoding in Fast-dLLM, PV-Decoding consistently improves sequential-MFP efficiency. In particular, PV-Decoding reduces the number of model forward passes by approximately 8\%--37\% relative to confidence-based decoding, with PV-1 providing the largest step reduction. Even PV-2, despite its stricter all-or-nothing verification rule, still reduces model forward passes by 7\%--22\% relative to the same baseline.

The relative decoding efficiency in Figure~\ref{fig:pareto-scatter} also reflects the expected ordering among PV-Decoding variants: overall, PV-1 achieves the highest step-based efficiency, followed by PV-multi and then PV-2. This trend is consistent with their verification policies. PV-1 verifies all draft tokens within a single branch and therefore minimizes verification overhead; PV-2 uses an all-or-nothing comparison between a baseline branch and a speculative branch, which is more conservative and yields smaller step reductions; and PV-multi verifies nested subsets of draft tokens, allowing partial speculative progress when the full speculative branch is not accepted and thus typically placing its efficiency between the other two variants.

In terms of accuracy, PV-Decoding exhibits a clear trade-off governed by verification cost. 
Stronger verification generally improves stability, but small differences preclude a strict ordering.
On GSM8K, all three variants achieve accuracy comparable to or slightly better than confidence-based decoding. On the programming benchmarks MBPP and HumanEval, PV-1 is less stable across models,
with the largest observed $\Delta$Acc drop relative to confidence-based decoding being 1.3\% on openPangu-7B-Diffusion-Base for HumanEval. 
PV-2 and PV-multi generally narrow this gap by applying stricter verification: the PV-2 loss is below 1\%, while PV-multi generally best preserves accuracy relative to Fast-dLLM.
Overall, the results support the intended design of PV-Decoding: lighter verification yields larger step reduction, whereas stronger verification improves robustness.

This progressive verification design also makes PV-Decoding adaptable to different deployment scenarios. 
For tasks where small speculative deviations are less costly, 
such as GSM8K-style mathematical reasoning, 
PV-1 is often attractive because it incurs the lowest verification cost and provides the largest reduction in model forward passes while maintaining comparable accuracy. 
For accuracy-sensitive settings, including code generation and agentic workflows where a single incorrect token may propagate into execution failures or invalid tool calls, 
PV-2 or PV-multi provides a more conservative choice. 
These variants spend additional verification computation to reduce speculative errors, making them better suited when robustness is prioritized over maximum step reduction.

\subsection{Wall-Clock Performance}

We evaluate wall-clock performance from a fixed-resource deployment perspective. Table~\ref{tab:wall-clock} measures Fast-dLLM and all PV-Decoding variants in the same SGLang runtime on Dream-7B-Instruct/GSM8K, using one Ascend NPU at request concurrencies 1, 4, and 8. 
Holding the hardware budget fixed gives a direct comparison of the speedups available to a user without provisioning additional accelerators. 
It is nevertheless a deliberately resource-constrained view: PV-2 and PV-multi trade additional parallel branch computation for fewer sequential MFPs, so a single shared NPU---especially at high concurrency---cannot necessarily expose all of their available branch-level parallelism.

\begin{table*}[t]
    \centering
    \caption{Fixed-hardware performance on Dream-7B-Instruct/GSM8K using one Ascend NPU. Parentheses show changes from Fast-dLLM under the same concurrency.}
    \label{tab:wall-clock}
    \resizebox{\textwidth}{!}{%
    \begin{tabular}{c@{\hspace{8pt}}rrrr@{\hspace{14pt}}rrrr}
        \toprule
        & \multicolumn{4}{c}{Throughput (tok/s) $\uparrow$}
        & \multicolumn{4}{c}{Total evaluation time (s) $\downarrow$} \\
        \cmidrule(lr){2-5}\cmidrule(lr){6-9}
        Concurrency & Fast-dLLM & \shortstack{PV-Decoding\\(1-batch)} & \shortstack{PV-Decoding\\(2-batch)} & \shortstack{PV-Decoding\\(multi-batch)}
        & Fast-dLLM & \shortstack{PV-Decoding\\(1-batch)} & \shortstack{PV-Decoding\\(2-batch)} & \shortstack{PV-Decoding\\(multi-batch)} \\
        \midrule
        1 & 183.8 & 208.7 (+13.6\%) & 204.9 (+11.5\%) & 134.8 ($-$26.6\%)
          & 338.29 & 294.73 ($-$12.9\%) & 300.85 ($-$11.1\%) & 458.51 (+35.5\%) \\
        4 & 280.6 & 306.3 (+9.1\%) & 303.0 (+8.0\%) & 155.0 ($-$44.8\%)
          & 220.22 & 199.41 ($-$9.5\%) & 201.86 ($-$8.3\%) & 395.36 (+79.5\%) \\
        8 & 311.5 & 341.4 (+9.6\%) & 333.3 (+7.0\%) & 116.5 ($-$62.6\%)
          & 197.26 & 179.01 ($-$9.3\%) & 182.91 ($-$7.3\%) & 524.30 (+165.8\%) \\
        \bottomrule
    \end{tabular}%
    }
\end{table*}

Under this fixed hardware budget, PV-1 and PV-2 consistently improve throughput by 7.0\%--13.6\% and reduce total evaluation time by 7.3\%--12.9\% across all tested concurrencies. 
PV-2 is slightly slower than PV-1, consistent with the additional parallel work introduced by its second verification branch; nevertheless, its positive end-to-end speedup shows that the reduction in sequential decoding rounds already outweighs this overhead. 
PV-multi does not reach this crossover point in the current system: its larger set of verification branches reduces sequential MFPs but lowers throughput by 26.6\%--62.6\% and increases evaluation time by 35.5\%--165.8\%. Thus, the fixed-budget results establish practical acceleration for PV-1 and PV-2, but not for PV-multi.

These measurements reflect our current implementation rather than the full systems potential of branch-parallel verification. 
We implement each verification branch as an ordinary batch element, which incurs batching overhead, repeated memory and attention work across branches under the limited NPU capacity. 
This simple implementation leaves substantial optimization space analogous to tree-structured speculative verification: branches share a common context with potentially reusable states. 
Packing the branches into a tree-structured attention layout and verifying them with fused operators instead of independent batched forwards~\cite{miao2024specinfer} can maximize the performance of multi-branch verification under fixed hardware budget.
Such specialized kernels are particularly important for realizing PV-2 and PV-multi speedups; additional hardware parallelism is another possible operating point but is not evaluated here. 
We therefore interpret Table~\ref{tab:wall-clock} together with the sequential-MFP results: the former reports speedups realized by the present single-hardware implementation, while the latter quantifies algorithmic depth reduction whose conversion into further wall-clock gains depends on more efficient branch execution.

\subsection{Comparison with Prior Methods}

In the model-averaged comparison with prior methods, PV-Decoding is competitive in accuracy versus sequential MFPs.
LocalLeap achieves relatively large reductions in MFPs.   
However, these reductions often come with substantial accuracy degradation, especially on GSM8K and HumanEval. 
LoPA exhibits unstable accuracy across settings and causes accuracy drops exceeding 5\% in some cases. 
Spiffy is competitive in selected scenarios, but it relies on multiple verification branches to verify multiple draft tokens, so its step reduction is closely tied to the branch budget $k$; increasing $k$ can expose more speculative progress, but also requires more verification computation. Moreover, its relative decoding efficiency is less stable across models and benchmarks.

Across most settings, PV-2 and PV-multi retain sequential-MFP reductions while improving robustness in task accuracy. PV-1 incurs no extra verification branch, PV-2 requires only one extra branch, and PV-multi increases branch count only as needed.

\section{Discussion}
\begin{figure*}[htbp]
    \centering
    \includegraphics[width=0.95\linewidth]{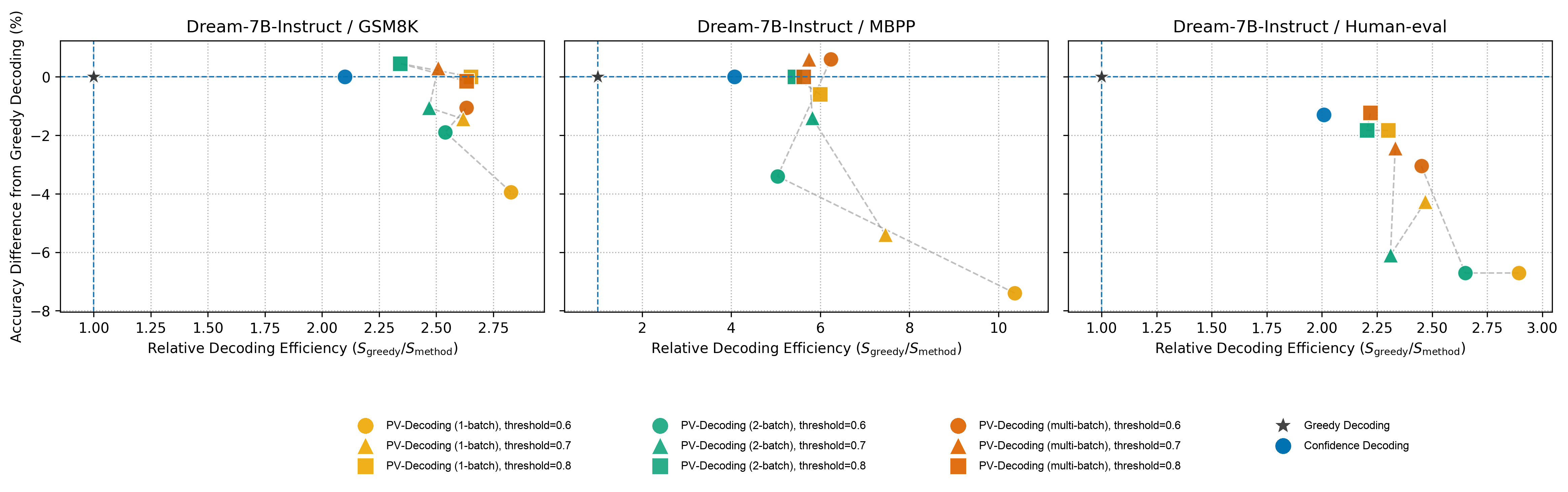}
    \caption{Accuracy--efficiency trade-off on Dream-7B-Instruct under different speculative confidence thresholds $\tau_{\mathrm{spec}} \in \{0.6, 0.7, 0.8\}$.}
    \label{fig:ablation-spec-confidence}
\end{figure*}

\subsection{Ablation on the Speculative Confidence Threshold}

Figure~\ref{fig:ablation-spec-confidence} examines the speculative confidence threshold $\tau_{\mathrm{spec}}$ on Dream-7B-Instruct with $\tau_{\mathrm{dec}}=0.9$. Lower thresholds enlarge the speculative set and reduce diffusion decoding steps, but also make speculative tokens less reliable, especially on programming benchmarks.

The figure shows a consistent accuracy--efficiency trade-off. When $\tau_{\mathrm{spec}}=0.6$ or $0.7$, PV-Decoding often reaches higher relative decoding efficiency but suffers larger $\Delta$Acc drops. Setting $\tau_{\mathrm{spec}}=0.8$ yields a more conservative policy: it sacrifices part of the step reduction while keeping $\Delta$Acc close to confidence-based decoding. Stronger verification further mitigates errors caused by aggressive speculation, as dual-batch and multi-batch verification generally improve accuracy over single-batch verification at the same threshold.
All main experiments use $(\tau_{\mathrm{dec}},\tau_{\mathrm{spec}})=(0.9,0.8)$ without task-specific tuning; the thresholds are fixed, but the selected token set varies by input and step.



\subsection{Ablation on Verification-Gated Decoding}

We further ablate the verification-gated decoding rule in single-batch verification. This rule commits new masked positions with confidence above $\tau_{\mathrm{dec}}$ only when all speculative tokens from the previous step pass verification; otherwise, the current step avoids committing additional direct tokens on an unreliable speculative context.

\begin{table}[t]
\centering
\caption{Ablation of verification-gated decoding in PV-Decoding (1-batch). ``w/o decode gating'' allows new high-confidence positions to be committed even when some speculative tokens fail verification.}
\label{tab:decoding_results}
\resizebox{\columnwidth}{!}{%
\begin{tabular}{llcc}
\toprule
Dataset & Method & Acc & MFP  \\
\midrule
\multirow{3}{*}{GSM8K}
& Fast-dLLM          & 72.4 & 58.3  \\
& PV-Decoding (1-batch)               & 72.4 & 46.2  \\
& PV-Decoding (1-batch) w/o decode gating & 70.6 & 49.9  \\
\midrule
\multirow{3}{*}{HumanEval}
& Fast-dLLM          & 54.8 & 86.4  \\
& PV-Decoding (1-batch)               & 54.3 & 75.5  \\
& PV-Decoding (1-batch) w/o decode gating & 50.0 & 70.8  \\
\midrule
\multirow{3}{*}{MBPP}
& Fast-dLLM          & 55.4 & 62.9  \\
& PV-Decoding (1-batch)               & 54.8 & 42.7  \\
& PV-Decoding (1-batch) w/o decode gating & 51.3 & 35.8  \\
\bottomrule
\end{tabular}%
}
\end{table}

Table~\ref{tab:decoding_results} shows that verification-gated decoding primarily serves as a robustness mechanism for the low-overhead PV-1 variant. Decode gating improves accuracy on GSM8K, HumanEval, and MBPP, indicating that suppressing new direct commits after failed verification reduces error propagation from incorrect speculative tokens.

\section{Conclusion}

We presented PV-Decoding, a confidence-driven speculative decoding framework for discrete diffusion language models. PV-Decoding extends confidence-based block decoding by tentatively filling medium-confidence tokens and verifying them through progressively stronger policies. This design makes the accuracy--efficiency trade-off explicit: single-batch verification offers the lowest-overhead acceleration, while dual-batch and multi-batch verification provide stronger safeguards against erroneous speculative commitments. Experiments across models and benchmarks show that PV-Decoding reduces model forward passes while maintaining competitive accuracy. Hardware-matched tests verify PV-1/PV-2 speedups; PV-multi requires optimized branch execution for wall-clock gains.

\section{Limitations}

PV-Decoding relies on confidence scores to select medium-confidence positions, so its effectiveness depends on model calibration and task sensitivity; adaptive speculative thresholds remain future work. Programming benchmarks also appear more vulnerable than GSM8K to incorrect speculative commitments, although a detailed error taxonomy is still needed. Finally, practical gains depend on efficient verification-branch batching. PV-1 and PV-2 add zero and one branch, respectively, and therefore have clearer paths to acceleration, whereas PV-multi incurs greater overhead and may require specialized kernels, efficient batching, or additional compute to translate its algorithmic gains into wall-clock speedups.

\bibliography{references}   
\end{document}